\documentclass[11pt,a4paper]{article}

\usepackage{ccaption}
\usepackage[utf8]{inputenc}
\usepackage{amsmath,amssymb}
\usepackage{setspace}
\usepackage{afterpage}
\usepackage{geometry}
\usepackage{graphicx}
\usepackage{sectsty}
\sectionfont{\fontsize{12}{15}\selectfont}
\usepackage{fancyhdr}
\usepackage{indentfirst}
\usepackage{multirow}

\usepackage{url}

\usepackage{soul} 

\usepackage{array}
\usepackage{color}
\definecolor{mygray}{gray}{0.6}

\newcolumntype{C}[1]{>{\centering\arraybackslash}p{#1}}
\usepackage[skip=0pt]{caption}

\usepackage{bibunits}
\usepackage{cite}
\defaultbibliography{main}
\defaultbibliographystyle{naturemag}

\usepackage{hyperref}
\hypersetup{
}

\makeatletter
    \def\hyper@natlinkstart#1{%
      \Hy@backout{#1}%
      \hyper@linkstart{cite}{cite.\@bibunitname.#1}%
      \def\hyper@nat@current{#1}%
    }
    
    \def\hyper@natlinkbreak#1#2{%
      \hyper@linkend#1\hyper@linkstart{cite}{cite.\@bibunitname.#2}%
    }
    
    \def\hyper@natanchorstart#1{%
      \hyper@anchorstart{cite.\@bibunitname.#1}%
    }
    
    \def\bibcite#1#2{%
      \@newl@bel{b}{#1}{\hyper@@link[cite]{}{cite.\@bibunitname.#1}{#2}}%
    }%
    
    \def\@lbibitem[#1]#2{%
      \@skiphyperreftrue
      \H@item[\hyper@anchorstart{cite.\@bibunitname.#2}%
      \@BIBLABEL{#1}\hyper@anchorend\hfill]%
      \@skiphyperreffalse
      \if@filesw
        \begingroup
          \let\protect\noexpand
          \immediate\write\@auxout{%
            \string\bibcite{#2}{#1}%
          }%
        \endgroup
      \fi
      \ignorespaces
    }%
    
    \def\@bibitem#1{%
      \@skiphyperreftrue\H@item\@skiphyperreffalse
      \hyper@anchorstart{cite.\@bibunitname.#1}\relax\hyper@anchorend
      \if@filesw
        \begingroup
          \let\protect\noexpand
          \immediate\write\@auxout{%
            \string\bibcite{#1}{\the\value{\@listctr}}%
          }%
        \endgroup
      \fi
      \ignorespaces
    }%
    
    \def\@citex[#1]#2{%
      \let\@citea\@empty
      \@cite{%
        \@for\@citeb:=#2\do{%
          \@citea
          \def\@citea{,\penalty\@m\ }%
          \edef\@citeb{\expandafter\@firstofone\@citeb}%
          \if@filesw
            \immediate\write\@auxout{\string\citation{\@citeb}}%
          \fi
          \@ifundefined{b@\@citeb}{%
            \mbox{\reset@font\bfseries ?}%
            \G@refundefinedtrue
            \@latex@warning{%
              Citation `\@citeb' on page \thepage \space undefined%
            }%
          }{%
            \hyper@natlinkstart{\@citeb}%
                \hbox{\csname b@\@citeb\endcsname}%
            \hyper@natlinkend%
          }%
        }%
      }{#1}%
    }%
\makeatother

\newgeometry{
    left=1.0in,
    right=1.0in,
    top=1in,
    bottom=1in
}
\usepackage[
    colorinlistoftodos,
    textwidth=2.1cm
]{todonotes}

 \usepackage{afterpage} 
 \newcommand\blankpage{%
    \null
    \newpage}

 \usepackage{xr}
 \makeatletter
 \newcommand*{\addFileDependency}[1]{SI.tex
   \typeout{(#1)}
   \@addtofilelist{#1}
   \IfFileExists{#1}{}{\typeout{No file #1.}}
 }
 \makeatother
 
\title{The Hierarchical Organization of Syntax}

\makeatletter
\renewcommand*{\@fnsymbol}[1]{\ensuremath{\ifcase#1\or \dagger\or *\or  \ddagger\or
    \mathsection\or \mathparagraph\or \|\or **\or \dagger\dagger
    \or \ddagger\ddagger \else\@ctrerr\fi}}
\makeatother    

\usepackage[symbol]{footmisc} 

\author{Babak Ravandi$^{1,2,\dagger,}$\footnote{Authors equally contributed.} \and Valentina Concu$^{3,*,}$\footnote{Corresponding authors e-mail: bk.ravandi@gmail.com and vconcu@uninorte.edu.co}
}

\date{%
    $^1${\small Network Science Institute, Northeastern University, Boston, USA}
    \vspace{-7pt}\\
    $^2${\small Department of Physics, Northeastern University, Boston, USA} 
    \vspace{-7pt}\\
    $^3${\small Department of Foreign Languages, Universidad del Norte, Barranquilla, Colombia}
    \vspace{-7pt}
    \\[2ex]%
}

\DeclareRobustCommand{\txr}[1]{{\textcolor{red}{#1}}} 

\usepackage{lineno}

\begin{document}

    


\maketitle


\vspace{1cm}

Hierarchies are the hidden backbones of complex systems and their analysis allows for a deeper understanding of their structure and how they evolve. 
We consider languages also to be complex adaptive systems with several intricate networks that capture their structure and function. 
Hence, we decided to analyze the hierarchical organization of historical syntactic networks to understand how syntax evolves over time. 
We created these networks from a corpus of German texts from the 11th to 17th centuries, focusing on the hierarchical levels of these networks. 
We developed a framework to empirically track the emergence of syntactic structures diachronically, enabling us to map the communicative needs of speakers with these structures. 
We named these syntactic structures ``syntactic communicative hierarchies.'' We showed that the communicative needs of speakers are the organizational force of syntax. 
Thus, we argue that the emergence of syntactic communicative hierarchies plays a crucial role in shaping syntax over time. 
This may indicate that languages evolve not only to increase the efficiency of transferring information, but also to increase our capacity, as a species, to communicate our needs with more and more sophisticated abstractions.

\blankpage

\section*{Introduction}


Among all the known means of communication, the human language is the only one that developed syntax \cite{Animal_communication,evolution_of_communication,Language_and_species}, which regulates how words combine into phrases and how phrases combine into larger phrases \cite{van1997syntax}. 
This regulatory strength enables speakers to make sense of the underlying combinatory features of word association. 
It also allows them to construct a rich repertoire of sentences from a set of words within the boundaries imposed by semantics and pragmatics \cite{wechsler2015word}, ``making infinite use of finite means'' \cite{aspects_of_the_theory_of_syntax}.  



The current shape of syntax in any given language is the result of evolutionary processes that have been triggered by the communicative needs that speakers needed to satisfy over centuries \cite{emergent_grammar}.
Hence, the syntactic rules by which we speak today can be defined as ``epiphenomenal'' of communication \cite{emergent_grammar, hopper201514,nettle1999linguistic,cameron2007complex}.
The necessity to communicate time references, for instance, has prompted many languages to develop an effective system of verbal tenses and temporal adverbs \cite{bybee1994evolution}. 
Such a system makes it possible for speakers to specify the time frame in which actions take place. 
In today's English, for instance, such a system counts on different morphologically marked forms and syntactic constructions (``I went'' vs ``I have gone'') that speakers use based on the meaning needed to be conveyed. 
In a similar fashion, different communicative needs among German speaking communities have resulted into syntactic changes and the emergence of new constructions. Fig. \ref{fig_communicative_need_and_syntactic_change} shows syntactic changes around three common communicative needs at different stages in the history of German \cite{leuschner2005grammatikalisierung}. 

While the role of communication in the evolution of syntax has been widely recognized by the linguistic community \cite{emergent_grammar,hopper201514,nettle1999linguistic,cameron2007complex}, we are still missing a framework that allows us to empirically track how speakers and their communicative needs shape syntax over time. This may also be due to the fact that quantitative methods in historical linguistic research are still scarce \cite{mcgillivray2023quantifying}.
Hence, in this study, we argue that the diachronic investigation of the syntactic hierarchical structures using complex  networks can provide a novel approach to understand how syntax evolves.
Our approach is motivated by two main reasons. 
First, we consider languages to be complex adaptive systems \cite{cameron2007complex, five_grace_language_complex_intro}. And, since ``behind every complex system there is an intricate network that encodes the interactions between the system’s components'' \cite{barabasi2013network_intro}, we decided to use the tools offered by network science \cite{amancio_perspective_language_complexity_intro,choudhury2009structure, small_world_language_intro, steyvers2005large, vitevitch2008can, cong2014approaching} for our investigation. 
Furthermore, studying languages as complex networks allows to focus on the linguistic interactions (e.g., nominal, verbal, and prepositional phrases represented as links) between multiple elements (e.g., phonemes, morphemes, or words represented as nodes), rather than analysing these elements separately  \cite{beckage2016language}. 
Second, hierarchies are recognized as an essential characteristic of complex systems \cite{ravasz2003hierarchical, luo2011detecting}. 
Thus, we introduce a framework that enables studying and tracking the emergence of hierarchies in diachronic syntactic networks, revealing how the elements of these networks relate to each other. 
The developed framework makes it possible to study and track the emergence of hierarchies in syntactic networks through multiple centuries, observing how the elements of these networks relate to each other. 
Specifically, we used historical syntactic networks from Middle and Early New High German (1050 - 1700) that were manually built following the principles of dependency grammar \cite{parsing2009speech}.
We followed the emergence of new syntactic constructions and their corresponding hierarchies in these networks, measuring their hierarchical levels and identifying the most influential nodes \cite{graph_hierarchy}.
Our aim is to provide a quantitative approach for the study of language change, and more specifically, to show how communication shapes syntax over time.

\section*{Results}

We studied the hierarchical structure of our syntactic networks from a diachronic prospective by building eight networks from a corpus of Middle and Early New High German texts (1050 – 1650) following the rules of dependency grammar \cite{parsing2009speech}. 
This means that we had to create first syntactic trees from the sentences we selected from the corpus and merge them together. 
We show how historical syntactic networks can be used to track the hierarchical evolution of syntax over centuries (see Methods).





To create the historical syntactic networks, we merged all the syntactic trees we built from our corpus.   
All the eight networks 
represent the sum of the sentence graphs found in the corpus. 
Hence, we refer to these networks as Aggregated Syntactic Networks, henceforth ASNs.
Fig. \ref{fig_ASN} shows how we built ASNs from dependency syntactic trees as well as our choice to separate words based on the syntactic role of words in a sentence. 
In every ASN, all the nodes, which correspond to the lemma forms of the words, are labeled according to their grammatical function (Table \ref{tbl_roles_rules}). 
Such a labeling allows to individuate the connective patterns of each node and how these patterns change over time. 
It also allows for analyzing which grammatical functions are the most influential or which ones are involved in the rise of new structures and their related hierarchies.

\subsubsection*{Characteristics of Aggregated Syntactic Networks}
After creating our eight ASNs, we studied their general characteristics: number of nodes, number of edges, average path length, and cluster coefficient. This information is presented in Table \ref{tbl_network_specs}. 
We also investigated the differences between the the dependency trees we created with our ASNs. To do that, we compared the depth of the trees with the lengths of the shortest distances in our ASNs. 
We observed a structural difference in our ASNs starting from the 16th century, as shown in Fig. \ref{fig_tree_depth_vs_network_diamet}.
The network diameter and average pair-wise shortest path length suddenly are as twice as the depth of largest sentence in our corpora. 
This indicate that the merged dependency trees mirror the syntactic changes in the global syntactic characteristics of the language as new periphrastic constructions emerged (such as the periphrastic future with \textit{werden} \cite{future_vale}) and are slowly used more frequently. 
One last characteristic we investigated was whether the power-law distribution was statistically plausible in our ASNs. 
To do that, we utilized the semi-parametric bootstrap approach combined with goodness-of-fit tests based on the Kolmogorov-Smirnov (KS) statistic \cite{power_law_dist_aaron}.  
Fig. \ref{fig_degree_dist_1} and Fig. \ref{fig_degree_dist_2} present the results. 
The calculated p-value for the estimated power-law fit is $>0.01$ (except for the 17th century). 
This value signals that the power-law model is favored over the alternatives by the LRT, presenting a strong correlation for having power-law degree distributions. 


\subsection*{Hierarchies and Communicative Needs}

To gain access to the hierarchical structure of our ASNs, we decided to focus on the networks' trophic levels, since their application in ecological systems and their corresponding food chains has successfully laid the foundations for analyzing hierarchical structures in graphs \cite{trophic_dynamic_aspect_ecology}. 
Thereafter, the generalization of trophic levels to complex networks paved the way to assess hierarchical characteristics of complex systems \cite{graph_hierarchy}. 


In this work, we decided to use a local metric, the forward hierarchical level \cite{graph_hierarchy}, as it mimics the top-down flow of our ASNs, which are all governed by heads that can be classified as vertices (since they do not have any in-neighbours). We show that the distribution of the forward hierarchical levels in ASNs captures syntactic changes, as shown in Figure S\ref{fig_hierarchical_levels_dist}. 
By conducting a historical comparison of the hierarchical levels, we observed the rise of new hierarchies which represented distinct communicative needs. 
Hence, we named these hierarchies syntactic communicative hierarchies. 
We discovered that the rise of these new hierarchies corresponded to specific syntactic changes in the language.  

One of the changes we were able to capture was the rise of the modal verb \textit{können} (\textit{can} in English) in Early New High German as it replaced \textit{mögen} (\textit{to like} in English) to express mental and physical abilities.
\textit{Mögen} was the most frequent modal verb in Old High German and one of the most frequent ones in Middle High German\cite{german_modal_verbs}. Until the last centuries of Middle High German (14th century), \textit {mögen} was used to express abilities as well as other communicative needs such as expressing preferences and obligations \cite{german_modal_verbs}. 
\textit{Können}, which in Old and Middle High German was not yet an integral part of the group of modal verbs, was used instead to express a small range of intellectual capacities only.
However, in the last two centuries of the Early New High German period, \textit{können} became the preferred way to express physical and mental abilities \cite{german_modal_verbs}. 

The transition of the communicative need of expressing physical and mental abilities from \textit{mögen} to \textit{können} was captured by the hierarchical analysis in our ASNs, as shown in Fig. \ref{fig_mogen_konnen}. 
The vertical position of nodes in Fig. \ref{fig_mogen_konnen} represents the forward hierarchical levels of the words in our corpus from the 14th to the 17th centuries. 
\textit {Mögen}, marked with green star, is constantly located at the top of hierarchies, together with other modal and auxiliary verbs. 
At the beginning of the Early New High German period, \textit {können}, marked with the white star, starts to become part of the of the group of modal verbs, when the semantic restrictions that bounded this verb to limited communicative contexts began to be slowly lifted \cite{german_modal_verbs}. 
Consequently, \textit {können} gradually started to replace \textit {mögen} in more and more instances, until, in the 16th century, it became the preferred way to express abilities, both intellectual and physical \cite{german_modal_verbs}. 
The use of ASNs allowed us to capture the exact moment in which the communicative hierarchy with \textit{können} emerged, contributing to the strengthening of the multi-hierarchical organization of syntax.
We believe that processes similar to this one is what shaped syntax in the first place as language was changing from a non-syntactical to a syntactic means of communication \cite{evolution_of_syntactic_communication}. 


It is also worth noticing that the use of ASNs goes beyond the traditional frequency based analysis for language change. With the rise of \textit{können} as modal verb to express mental and physical capacities in the last two centuries of Early New High German, its frequency starts to increase \cite{german_modal_verbs}. Such an increase can also be observed in our corpus as well. \textit {Mögen}, on the other hand, experiences a drop in frequency, as it is slowly being replaced by \textit{können} to express mental and physical capacities.
Nonetheless, the hierarchical analysis of the ASNs ranks \textit {mögen} as high as \textit{können} in the last two centuries of the Early New High German period. 
Since forward hierarchical levels rank the influence of a particular vertex in a graph \cite{graph_hierarchy}, the results of our analysis suggests that \textit {mögen} was still as influential as \textit{können} in the last two centuries of the Early New High German period, even if its frequency in our corpus drops significantly. This is probably due to the fact that, in the 16th century, \textit {mögen}, started to be used to express ``desire''  \cite{german_modal_verbs}, and thus to be used as the head of a new syntactic communicative hierarchy.
Traditionally, historical linguistic analyses rely on frequency to track the evolution of specific words or constructions \cite{hopper2003grammaticalization}. 
However, and without relying solely on frequency, we could identify as \textit {mögen} and \textit{können} as two of the most influential nodes in our ASNs and to related such an influence to specific communicative needs.
Furthermore, our approach was also able to capture other syntactic changes similar to the rise of \textit{können} as a modal verb and its corresponding hierarchies. 
We provide the list of the other hierarchies we identified in our corpora in Table \ref{tbl_communictive_needs}.

Lastly, our results   suggest that similar processes as the ones discussed in this study could have given rise to different linguistic hierarchies when the human language was changing from a non-syntactic to a syntactic communication system, shaping syntax over time \cite{aspects_of_the_theory_of_syntax}. 
The resulting multi-hierarchical structure, as the one we were able to observe in the current work, is a reflection of the language effectively managing its duty to serve the communicative needs of the speakers. 
How heads are selected to carry out specific communicative needs is likely triggered by a combination of semantic and pragmatic factors that created the right environment for these hierarchies to rise. 











\subsection*{Limitations}

Satisfying the communicative needs of speakers is an extremely complex problem. 
To carry out this task, human languages have developed a broad variety of resources ranging from phonology and morphology to syntax and semantics. 
The main focus of this study is on syntax and its interplay with semantics. Hence, our ASNs do not capture other linguistic aspects such as morphology or phonology.
The main reason to focus on syntax is its structural uniqueness and characteristics as well as the type of data that is available to conduct historical linguistic analysis.
%

Moreover, the list of syntactic communicative hierarchies that we conceptualized in this work is not finite and is bounded to our corpus selection to analyse language evolution and change.
Our corpus selection is a representative of the language around the verb \textit{werden}. 
The reasons behind this choice are explained in the Methods section. 
To identify a broader range of syntactic communicative hierarchies, ASNs on more languages and their evolution over time should be studied. 
Also, our study revealed a multi-hierarchical structures behind the organization of languages through time.  
This does not convey that all aspects of language also organize through hierarchies as well.

Lastly, we have to acknowledge that working with historical data is challenging, since the availability of corpora is not as high as it is for modern languages. 
Also, available algorithms to parse language syntax only work with modern languages, so we had to manually build the syntactic trees for each sentence. 

\section*{Discussion}


In this paper, we provided a framework to study syntactic changes by analyzing the hierarchical structure of historical syntactic networks. 
Our approach was guided by two basic principles. 
First, we consider languages as complex systems that change and evolve to satisfy the communicative needs of the speakers over time \cite{five_grace_language_complex_intro,cameron2007complex,ellis2009language}.  
Second, we consider hierarchies as the invisible backbones of complex systems and hence languages \cite{Hierarchy_Theory_An_Overview,origins_hierarchy_complex_networks,Hierarchical_Organization_Modularity_Metabolic_Networks,Hierarchy_Measures_Complex_Networks}. 
The proposed approach enables connecting the communicative needs of speakers -- a derivative of the well studied role of communication in language change \cite{hopper201514,nettle1999linguistic,cameron2007complex} -- to specific syntactic hierarchies in ASNs. 
This framework enabled us 
to reveal hidden patterns that are the byproduct of the joint action of semantics and syntax. 
We named these newly discovered patterns ``syntactic communicative hierarchies,'' and we tracked their evolution through a time span of eight centuries.  
Doing that, we were able to empirically show how the interplay between communicative needs of the speakers and the act of communicating itself shape language. 
In our networks, this shaping action translated into the emergence of new syntactic hierarchies. Hence, our work provides empirical evidence to the claim that the grammar by which we speak today is ``epiphenomenal of communication'' \cite{emergent_grammar}. 





Our approach also allowed us to address a central question that arose during our analysis: what are the underlying mechanisms behind the emergence of those syntactic communicative hierarchies that we were able to identify from the hidden hierarchical characteristics of ASNs? 
The clues to answer this question lay in the global characteristics of the ASNs and their relation to the socio-individual needs of speakers to communicate with each other. 
Hence, we propose that the emergence of syntactic communicative hierarchies is a crucial driving force in what shapes syntax. 


We also would like to add that the ever-changing multi-hierarchical shape of syntax could also provide a better way to understand the emergence of Zipf's law in human language. 
Current theories on why syntactic communication was preferred by evolution are based on analyzing the patterns of growth in the number of actions/signals and objects in language  \cite{evolution_of_syntactic_communication,consequences_zipfs_law}. 
Nowak et al. suggest that the crucial step that guided the transition from a non-syntactic to a syntactic communication was an increase in the number of ``relevant events'' to which speakers needed to refer \cite{evolution_of_syntactic_communication}.  
Sole \cite{sole_syntax_free}, on the other hand, argues that the prerequisite for syntax is one of the fundamental principals of organizations that is common to all languages, the Zipf's law \cite{Least_effort_origins_scaling_human_language}. 
According to Solé \cite{sole_syntax_free}, is the Zipf's law what provides the ideal conditions for words to connect ``for free'' and is, therefore, an essential precondition for the language transition into a syntactic means of communication.  
We believe that our approach provides an alternative account to Sole's claim as it demonstrates how communication (and not the Zipf´s law) is responsible for for both the current structure and the origin of syntax.

We would like to conclude with the following: It is often considered that the general objective of language evolution is to maximize information transfer while minimizing the effort for the hearer \cite{Least_effort_origins_scaling_human_language}. 
However, the observed multi-hierarchical structure could provide a more precise explanation on the general goal of language being optimized through evolution. 
The human language is the battleground of communicating an ever-widening range of abstractions. 
The current communicative needs that we found in ASNs are the manifestation of the need to communicate abstract ideas that already found their way into the structure of our language.
The more advanced humans get, the more complex abstractions will need to be communicated. 
This means that the language will experience an increase in syntactic communicative hierarchies, which means a more multi-hierarchical organization. 

The multi-hierarchical structure of syntax could also be described as a polyarchy. 
The term polyarchy was coined in political science to differentiate between democracy as an ideal and its realization in human societies through processes like democratization \cite{polyarchy_dahl_book}. 
The concept of polyarchy and the processes behind its formation indicate a connection between the hierarchical organization of syntax and the emergence of polyarchy. 
We explored this possible connection between linguistics and political science in Section S\ref{SI_sec_possible_connection_with_Zipfs}.
As a future work, such explorations could be interpreted as an attempt to find a way to scientifically predict the trajectory of language change through centuries.

\section*{Methods}

\subsection*{Data Collection}


The texts from which we have collected the data come from two different database, one for Middle High German (Referenzkorpus Mittlehochdeutsch)\cite{klein2016referenzkorpus} and one for Early New High German (Bonner Frühneuhochdeutschkorpus)\cite{schmitz2011bonner}.
The selection of the texts was based on both internal and external textual features \cite{cavaglia2002measuring}. 
As for the first type, these features were based on syntactic, lexical, and semantics aspects. 
For instance, since poetic metric has been shown to affect and, sometimes, even ``distort syntactic structures,'' \cite{somers2014intersection} we selected prose texts only. 
Further, in order for us to concentrate on texts with similar topics, we limited our selection to religious, legal, and literary works.
The aforementioned criteria (text type and genre - syntactic, lexical, and semantic aspects) allow us to ensure ``corpus homogeneity'' \cite{cavaglia2002measuring} and make the results from Middle and Early New High German texts comparable with each other. 
The external features that influenced our selection were chronological and geographical origin. 
To work on a variety of texts that could best represent the language periods we are studying, we selected texts from all the attested five different dialect areas and from all the centuries traditionally included in the Middle and Early New High German periods.
Specifically, the Middle High German corpus consists of 30 different texts form the 11th, 12th–13th, and early 14th centuries in West Middle German, East Middle German; West Upper German, East Upper German, and North Upper German. 
In the same way, the corpus for Early New High German includes 20 prose texts from the late 14th, 15th, 16th, and 17th centuries in West Middle German, East Middle German, West Upper German, East Upper German, and North Upper German. 
This corpus was already used to track the development of \textit {werden} as the auxiliary for the expression of future references \cite{future_vale}.

Our corpus selection is a representative of the language around the verb \textit{werden}. \textit{Werden} can be combined with a large variety of elements depending on the grammatical functions it is carrying out in a sentence. 
The functions of \textit{werden} in Modern German are represented in Figure S\ref{fig_werden_in_modern_german}, showing how versatile this verb is in today's Modern German functioning as both a full verb and an auxiliary. From a historical perspective, werden acquired the capacity of carrying out such syntactic functions throughout the process of \textit{Desemantisierung} (desemantization) that took place starting in the Old High German period (9th and 10th centuries) and culminating in the Early New High German period (half of the 14th century -  17th century) \cite{kotin_werden}. Such processes caused, for instance, the increase of the use of \textit{werden} in the passive constructions. This process is also directly involved in the emergence of \textit{werden} as a future marker. This use emerged probably only in the last decades of the 14th century in the East Middle German and Upper High German dialect areas”  after rarely occurring with an infinitive verb in the Old High German period \cite{diewald_markers}.
The singularities of the development of \textit{werden} and it versatility make it, we believe, an exceptional candidate for our analysis. Focusing on \textit{werden} for both Middle and Early New High German also further ensures "corpus homogeneity" \cite{cavaglia2002measuring} and strengthen the comparability of our texts.  

We created two different databases, one for Middle High German and one for Early New High German. 
All the words are listed in their lemma form.
Each database contains all the tokens of \textit {werden} in the following combinations: [WERDEN + past participle], [WERDEN + present participle], [WERDEN + adjective], [WERDEN + infinitives], [WERDEN + nouns], and [modal verb + WERDEN]. 
Lastly, in the corpus there are missing annotations of those words that were missing from the original texts or of which the meaning is still unknown.  We decided to not include sentences with any missing element in the analysis that would not allow us to identify the function of \textit {werden}. The corpora display missing words annotated as [!] in the Middle High German corpus, and as ``unbekannt'' (unknown) in the Early New High German corpus. 
Figure S\ref{fig_werden_corpus_annotations} provides two examples of such a selection process. On the contrary, instances with missing annotations that did not interfered with the identification of the function of \textit{werden} were included.


The data was collected from the aforementioned annotated online corpora and the instances of \textit {werden} were manually parsed. 
This process is highly demanding in terms of time and limits the quantity of data that can be included for analysis. 
However, manually parsing allows us to have homogeneity across the corpora, which makes the comparison between Middle and Early New High German consistent.

\subsection*{Syntactic Dependency Trees} 

To create historical syntactic networks, we had to build first syntactic trees from a set of target sentences which were taken from a corpus of Middle (1050 – 1350) and Early New High German (1350 – 1650) texts. 
The corpus contained a total of 455,944 words distributed in 8 different centuries. 

Syntactic trees are commonly used in dependency grammar to map the relationships between words. Dependency grammar is based on the notion of ``grammatical relation'' according to which the relationship between a head and a dependent are established in a given sentence\cite{parsing2009speech}. 
The head is the word in the phrase that is grammatically the most important since it determines the syntactic category of of a larger constituent and its the central organizing word (e.g., the primary noun in a noun phrase, or verb in a verbal phrase). 
The remaining words in the constituent are either directly, or indirectly, dependent on their head.\cite{parsing2009speech}
Furthermore, ``the head-dependent relationship is made explicit by directly linking heads to the words that are immediately dependent on them''\cite{parsing2009speech}. 
The relationships established between the head and its dependent ``allows us to further classify the kind of grammatical relations, or grammatical functions, in terms of the role that the dependent plays with respect to its head.\cite{parsing2009speech} 
Dependency grammar uses ``dependency trees'' to structure and visualize the relationship between a head and its dependents. 
Figure S\ref{fig_syntactic_tree} shows a syntactic tree for a sentence in Modern English. 
According to Jurafsky and Martin,\cite{parsing2009speech} dependency structures as the one shown in Figure S\ref{fig_syntactic_tree} are direct graphs that respect the following constraints:

\begin{enumerate}
    
    \item There is a single designated root node that has no incoming arcs.

    \item With the exception of the root node, each vertex has exactly one incoming arc.

    \item There is a unique path from the root node to each vertex in the graph.

\end{enumerate}

\noindent We followed the aforementioned constraints to build individual syntactic dependency trees that were then merged to create syntactic networks for each of the century we included in this study.

\subsection*{Building Historic Syntactic Networks}

The rules used to build such networks are inspired by the principles of dependency grammar \cite{parsing2009speech}. 
There is a total of three rules in the current study, and these reflect the syntactic relations between the elements in a sentence. 
A rule is classified as a nominal phrase (NP) if it has a noun or a pronoun as the head. 
A rule is classified as a verbal phrase (VP) if it has a verb as the head. 
A rule is classified as a prepositional phrase (PP) if it has a preposition as the head. All the source and target nodes are listed in their lemma form which corresponds to the canonical word form (for instance, the singular form of a noun is the lemma for a plural form, or the infinitive form of a verb is the lemma of an inflected form of a verb). 
Each node in the network represents the lemma of a given word. We have also assigned to every node in the source and target columns a label according to its grammatical role. The complete list of grammatical roles is displayed in Table \ref{tbl_roles_rules}.

After assigning to each word its grammatical role, we created a syntactic tree for each sentence in which the target verb \textit {werden} was found. 
Afterwards, we merged all the sentence trees into one network for each one of the eight centuries included in our analysis. 
The resulting networks, which we decided to name Aggregated Syntactic Networks (ASN), are not trees anymore, as multiple paths now exist between heads and dependents.

\subsection*{Identifying Syntactic Communicative Hierarchies}

Communication has been recognized as a driving force for language evolution and change, and the grammar and the syntax of a language have been described as ``epiphenomenal'' of communication \cite{emergent_grammar}. 
In our work, communication also plays an important role. 
More specifically, we focused on what we called ``syntactic communicative hierarchies,'' which allowed us to take advantages of the interplay between semantics and syntax.
In our work, a communicative need corresponds to the resulting meaning expressed by the syntactic head and its dependants in any given tree. 
For instance, the trees with the auxiliary \textit{werden} as the syntactic head that has a past participle as a direct dependent is used express the passive voice. It serves the communicative need of focusing on the patient and not on the active subject of the sentence.
Hence, we used the syntactic heads to track down what kind of meanings could be mapped to specific hierarchies in our ASNs in a time spam of eight centuries.

Table \ref{tbl_communictive_needs} contains the list of syntactic communicative hierarchies that we identified in our ASNs. 
The chart shows the related syntactic heads, together with the historical stage, Middle High German (MHG) and Early New High German (ENHG) in which they were found or emerged. 
Note that some heads are associated with multiple communicative needs, while some of them acquired a communicative needs later on, like in the case of future references (transition from \textit{sollen} to \textit{werden}) or the expression of mental and physical capabilities (transition from \textit{mögen} to \textit{können}).

\bibliographystyle{naturemag}
\bibliography{main}

\begin{figure}[h!]
    \centering
    \includegraphics[scale=0.43]{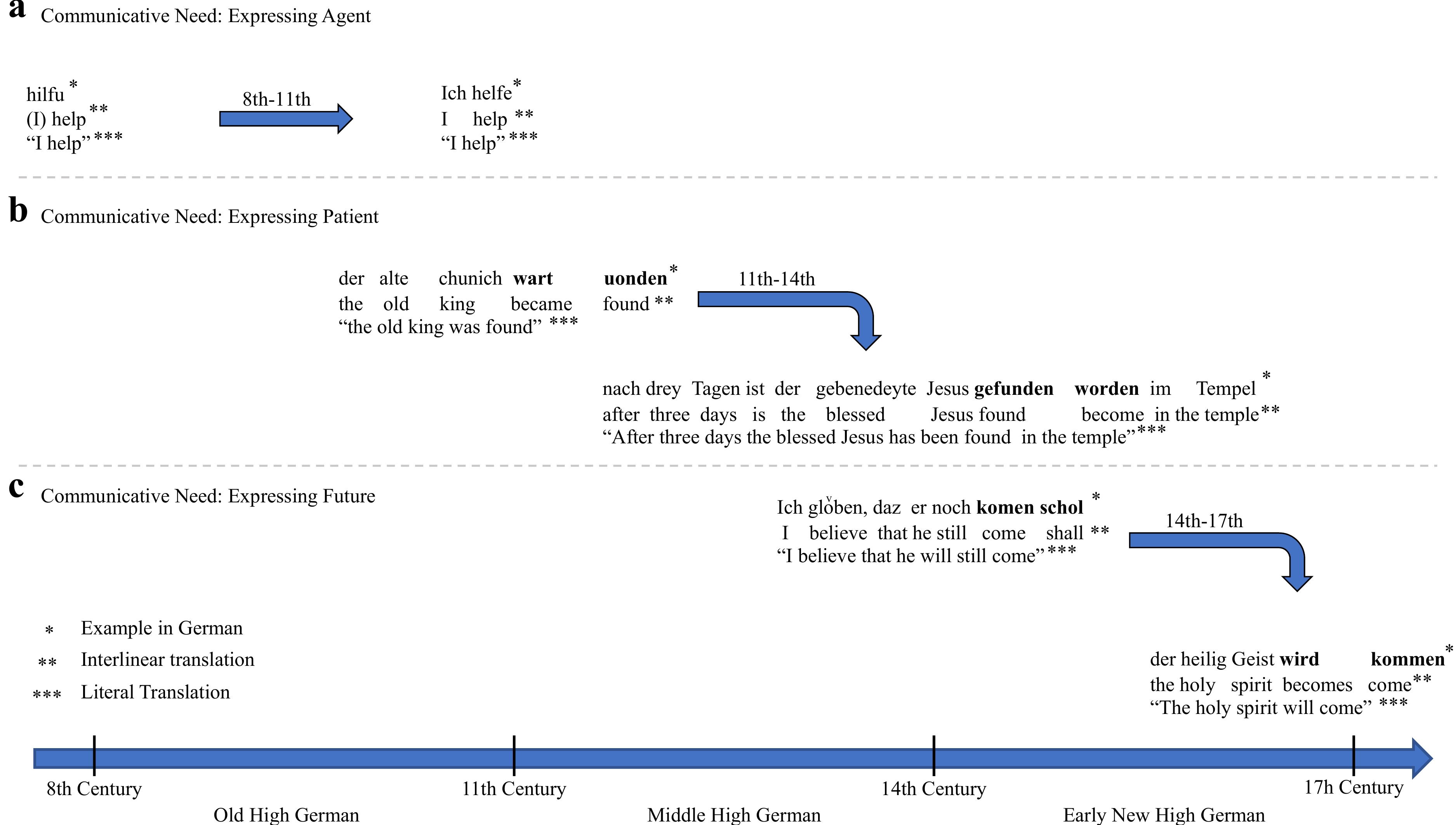}
    \vspace{10px}
    \caption{}
    \label{fig_communicative_need_and_syntactic_change}
\end{figure}
\begin{figure}[h!]
    \contcaption{
\textbf{The relationship between communicative needs and syntactic changes.} Communication has been often recognized to be a driving factor for language change in the syntactic inventory of a language. Hence, these changes always gravitate around specific communicative needs, they can be of distinctive types and affect one or more linguistic elements. \textbf{(A)} The verbs in Old High German were morphologically marked for the subject and no personal pronoun was needed to indicate who or what was doing a particular action. The loss of such morphological feature in Middle High German on one side, and the need to indicate the subject on the other, gives rise to the obligatory personal pronoun to accompany the main verb. \textbf{(B)} The emergence of the present perfect from Middle to Early New High German makes it possible for speakers to use the passive also in the newly grammaticalized tense. In this way, the communicative goal to emphasize the semantic role of the ``patient'' can be expressed in the past using two different tenses. 
\textbf{(C)} The rise of the constructions with the verb \textit{werden} and the infinitive verbs in Middle High German takes over a similar construction with the modal verb \textit{sollen} to satisfy the communicative need to express future reference in Early New High German.
    }
\end{figure}

\begin{figure}[h!]
    \centering
    \includegraphics[scale=0.30]{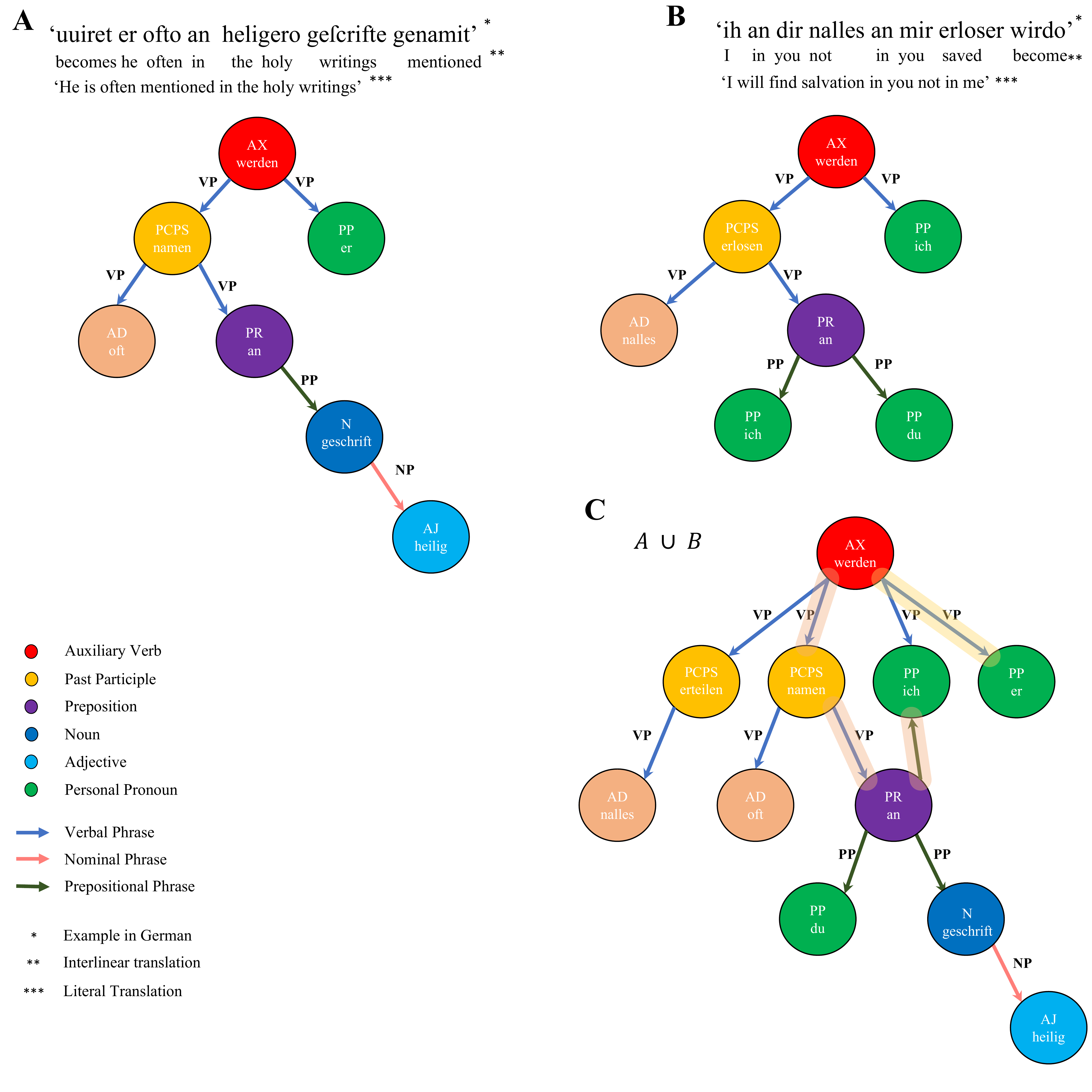}
    \caption{}
    \label{fig_ASN}
\end{figure}

\begin{figure}[h!]
    \contcaption{
\textbf{From Syntactic Dependency Trees to Aggregated Syntactic Networks.} 
A syntactic dependency tree represents the hierarchical structure of the constituents in any given sentence. 
However, to investigate the entire range of syntactic relationships in Middle and Early New High German, we aggregated the syntactic trees of all the sentences in which the target verb \textit{werden} was found. 
This approach enables us to shift focus from single sentences to a whole corpora during a particular century. 
Hence, an ASN is the representation of the sum of the syntactic relationships established between the sentences' constituents in a given century.
\textbf{(A-B)} Two syntactic dependency trees from the Middle High German corpus (11th century).
%
The labels of the nodes (presented here in distinctive colors) reflect the grammatical roles, whereas the edges refer to the syntactic relationships between the words.
\textbf{(C)} The resulting network by aggregating sentence A and B is no longer a tree because now two distinct paths exist from the root ``AX werden'' to the dependent ``PP er'', as highlighted with yellow and orange colors. 
Therefore, the aggregated network can not be treated as a dependency syntactic tree anymore. 
    }
\end{figure}

\begin{figure}[h!]
    \centering
    \includegraphics[width=\columnwidth]{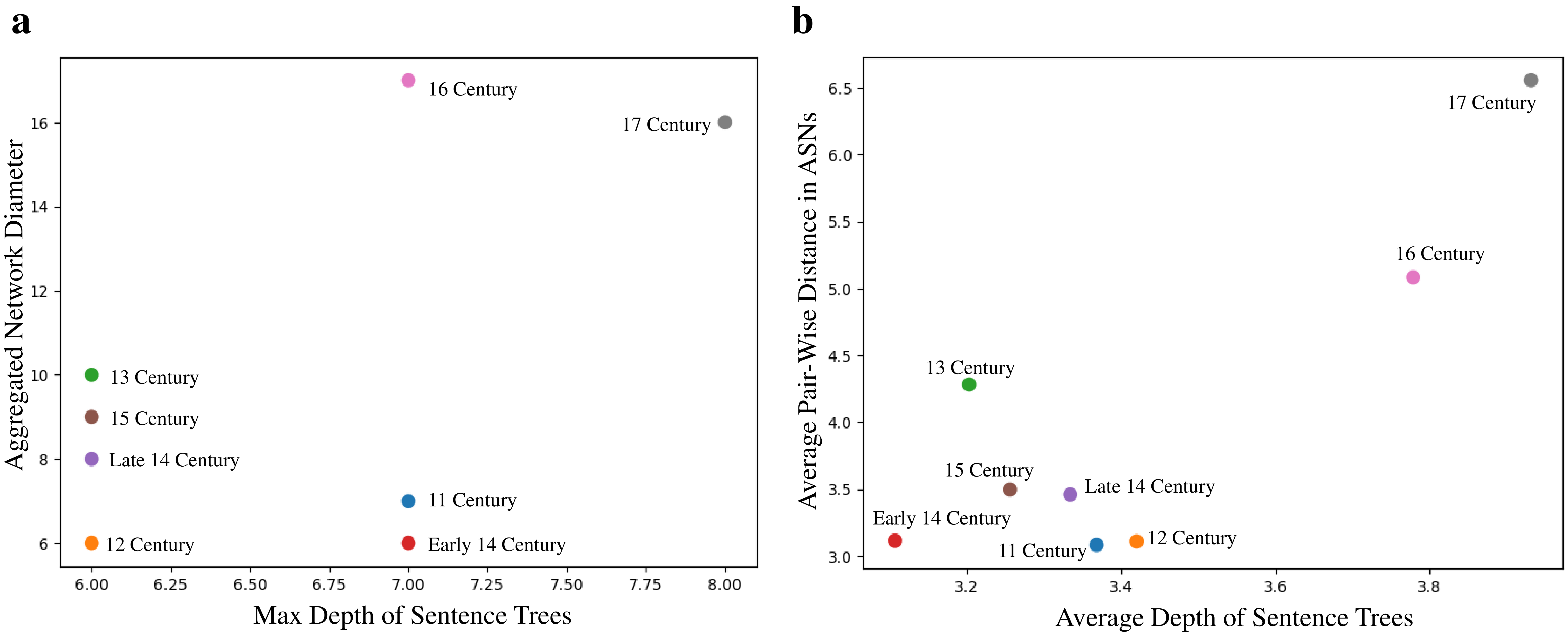}
    \vspace{10px}
    \caption{
    }
     \label{fig_tree_depth_vs_network_diamet}
\end{figure}
\begin{figure}[h!]
    \contcaption{
\textbf{Individual Tree's Depth vs Diameter of ASNs},
We compared the depth of the syntactic trees we created from our target sentences with the depth of our ASNs. 
While until the 16th century, the depth of both syntactic trees and ASNs is comparable, we observed a structural difference starting from the 16th century. 
The network diameter and average pair-wise shortest path length suddenly are as twice as the depth of largest syntactic tree in our corpora. 
This indicates that the merged syntactic dependency trees reflect the changes in the global syntactic characteristics of the language as new periphrastic constructions such as the future with \textit{werden} emerged and are slowly used more frequently \cite{future_vale}.
This phenomenon can be captured by our ASNs and will be investigated in further work.
    }
\end{figure}

\begin{figure}[h!]
    \centering
    \includegraphics[scale=0.40,page=1]{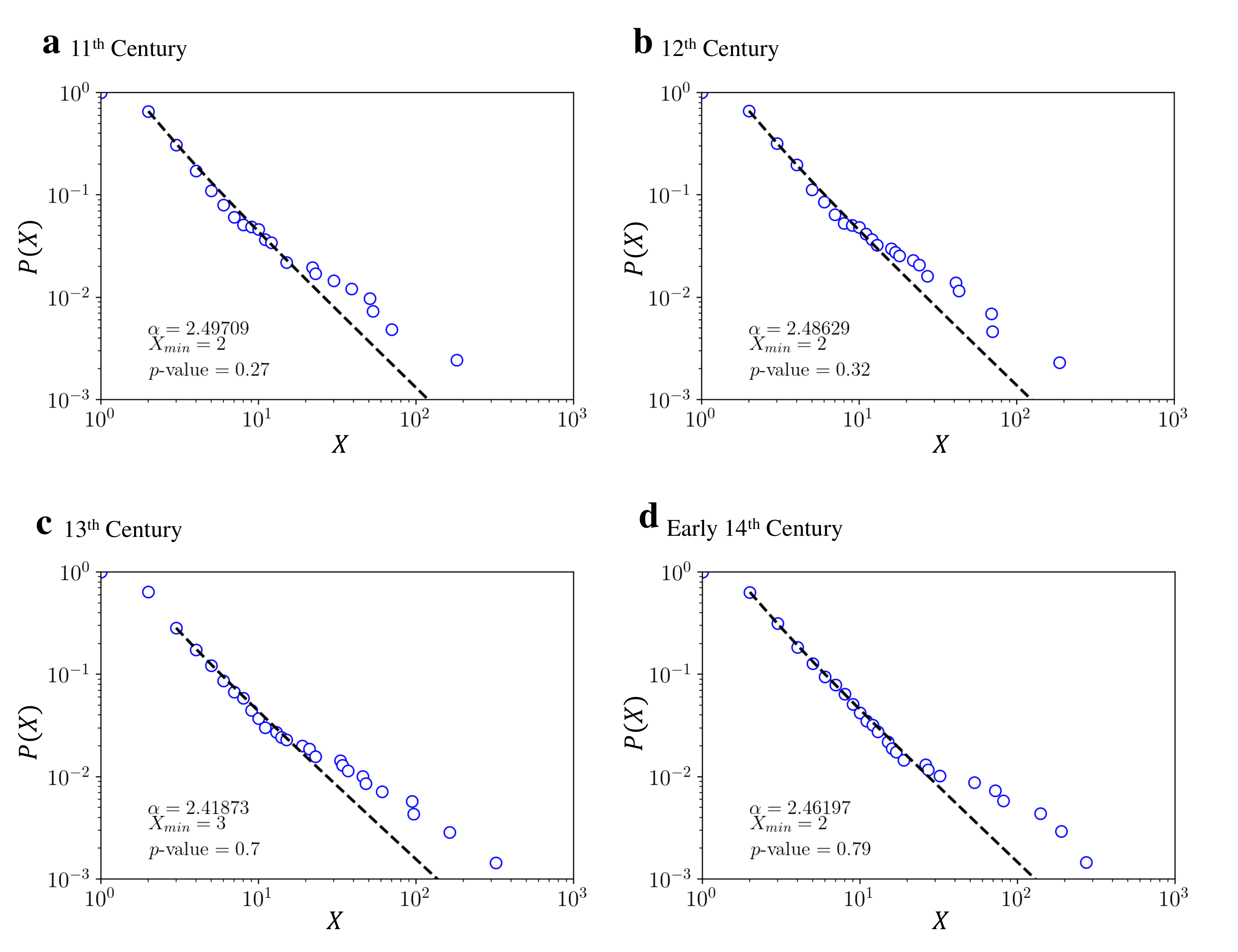}
    \caption{Degree distribution of ASNs in Middle High German.}
    \label{fig_degree_dist_1}
\end{figure}
\begin{figure}[h!]
    \contcaption{
\textbf{Degree distribution of ASNs in Middle High German}. 
The last characteristic we investigated
was whether the power-law distribution was statistically plausible in our ASNs. We carried out the analysis utilizing the semi-parametric bootstrap approach combined with goodness-of-fit tests based on the Kolmogorov-Smirnov (KS) statistic. The
calculated p-value for the estimated power-law fit is is $>0.01$.  
This value signals that the power-law model is favored over the alternatives by the LRT, presenting
a strong correlation for having power-law degree distributions in the Middle High German networks.    }
\end{figure}

\begin{figure}[h!]
    \centering
    \includegraphics[scale=0.40,page=2]{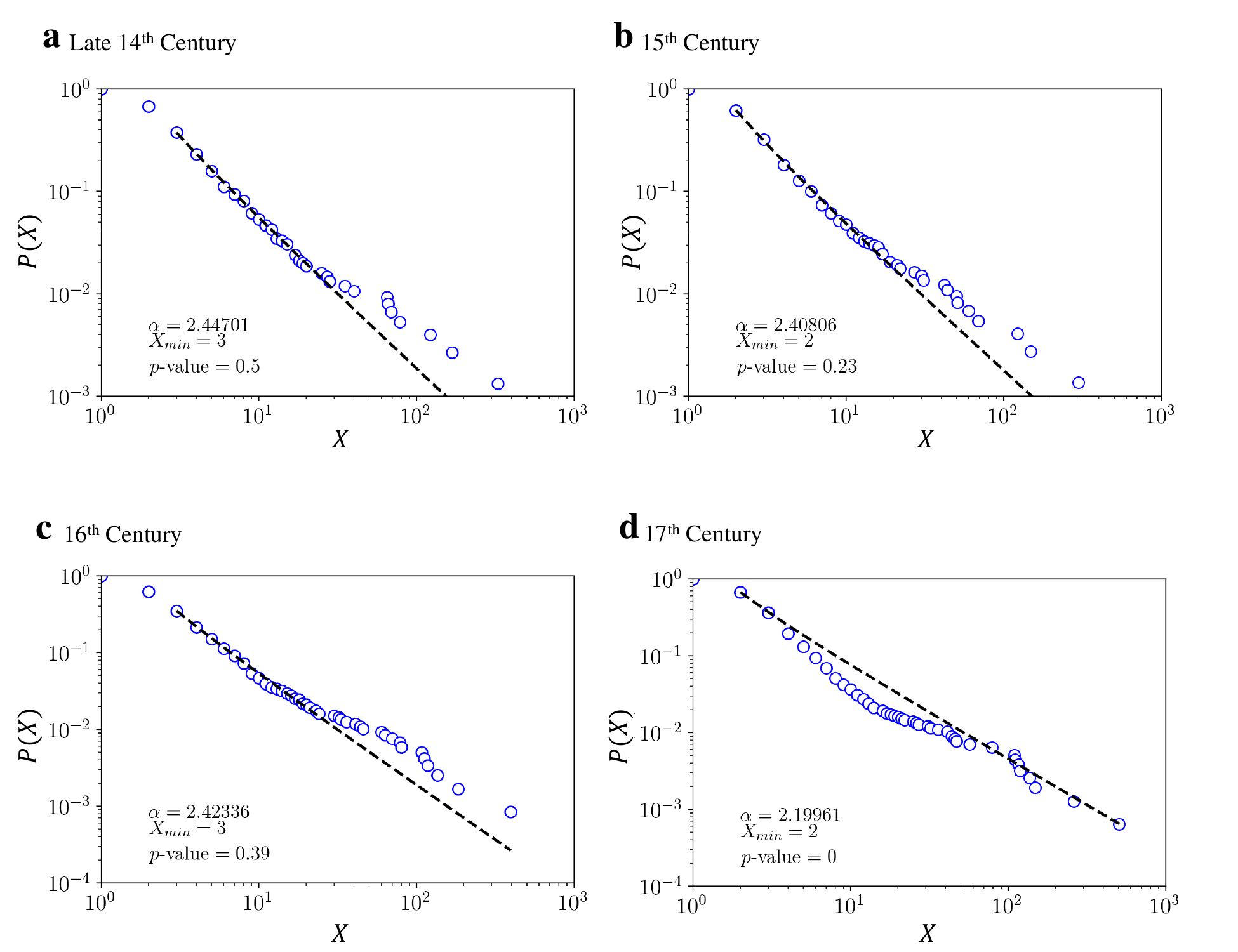}
    \caption{Degree distribution of ASNs in Early New High German.}
    \label{fig_degree_dist_2}
\end{figure}
\begin{figure}[h!]
    \contcaption{
\textbf{Degree distribution of ASNs in Early New High German}.
The calculated p-value for the estimated power-law fit is here too is $>0.01$ except for the 17th century. This
value signals that the power-law model is favored over the alternatives by the LRT for 3/4 of the Early New High German networks, ih which we found a
a strong correlation for having power-law degree distributions.
    }
\end{figure}

\begin{figure}[h!]
    \centering
    \includegraphics[width=\columnwidth]{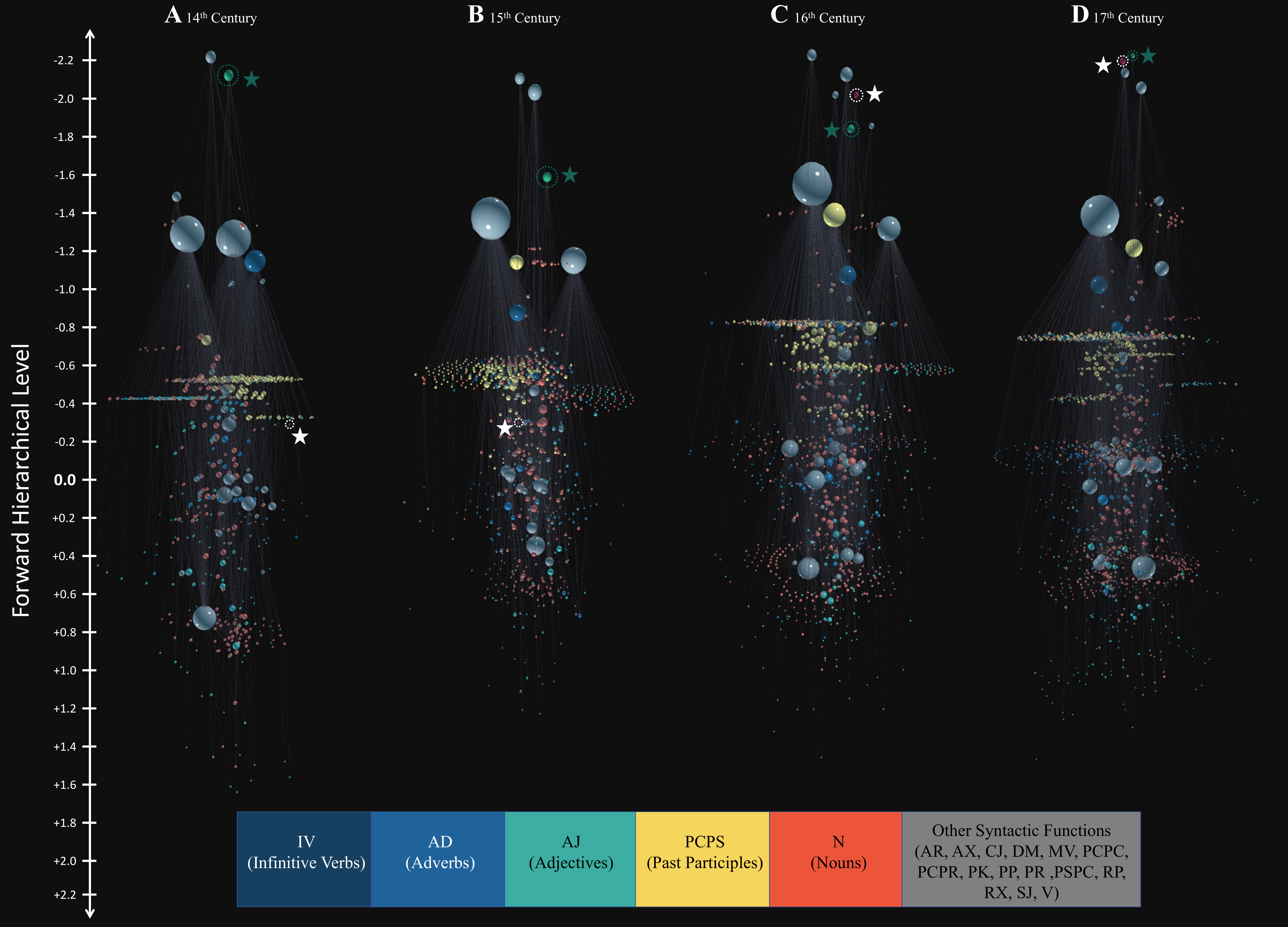}
    \vspace{10px}
    \caption{}
    \label{fig_mogen_konnen}
\end{figure}
\begin{figure}[h!]
    \contcaption{
\textbf{The Rise of \textit{können} as a Modal Verb to Express Physical and Mental Abilities.} 
The vertical positions of nodes represent the forward hierarchical levels in their respective ASNs (the 3D layout is generated using the algorithm in \cite{structural_transition}). 
The size of nodes indicates the number of syntactical connections a word has. 
\textit{Mögen} is marked with the green star and \textit{können} with the white star. 
\textbf{(a-b)} Through late Middle High German \textit{mögen} is used to express abilities as well as other communicative needs such as expressing preferences and obligations. 
\textbf{(c-d)} In the last centuries of the Early New High German period, a transition happened and \textit{können} became the preferred way to express physical and mental abilities. 
    }
\end{figure}


\blankpage\blankpage\blankpage\blankpage


\begin{table}[h!]
    \centering
    \caption{Grammatical Roles Used in ASNs.}
    \vspace{10px}
    \begin{tabular}{ p{2cm}p{5cm} |p{2cm}p{5cm} }
    \hline
    Label &Grammatical Role &Label &Grammatical Role \\
    \hline
    AD   & Adverb    &PR&   Preposition\\
    AJ&   Adjective  &PP   &Personal Pronoun\\
    AR &Article & PS&  Possessive Pronoun\\
    AX    &Auxiliary & PCPR&  Present Participle\\
    CJ & Coordinating Conjunction  & PCPS&Past Participle\\
    DM & Demonstrative Pronoun  & RX   &Reflexive Pronoun\\
    IV & Infinitive Verb  & RPO&Relative Pronoun\\
    MV & Modal Verb  & SC&Subordinating Conjunction\\
    N & Noun  & V & Verb\\
    PK & Particle &  &  \\
    \hline
    \end{tabular}
    \label{tbl_roles_rules}
\end{table}

\begin{table}[h!]
    \centering
    \caption{Characteristics of Aggregated Syntactic Networks}
    \includegraphics[scale=0.6]{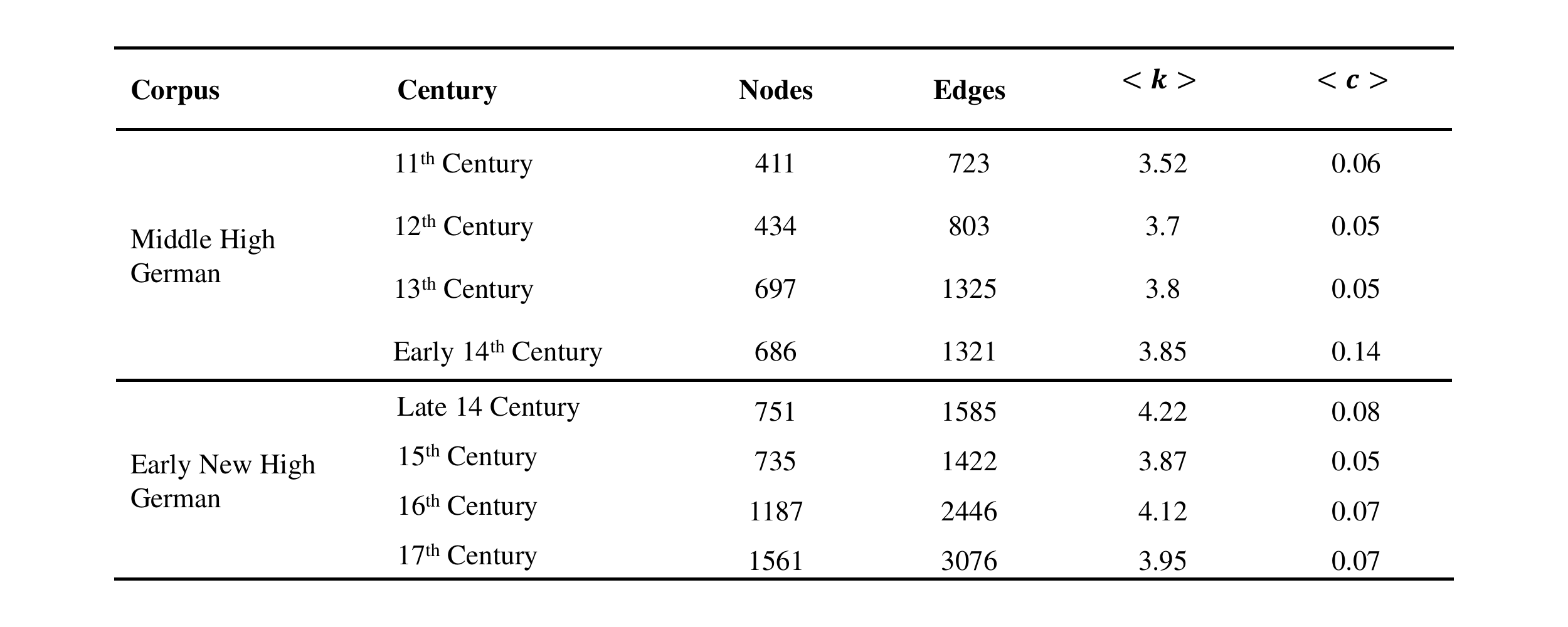}
    \label{tbl_network_specs}
\end{table}

\begin{table}[h!]
    \centering
    \caption{Syntactic Communicative Hierarchies in the Middle and Early New High German Corpus}
    \vspace{10px}
    \includegraphics[scale=0.7]{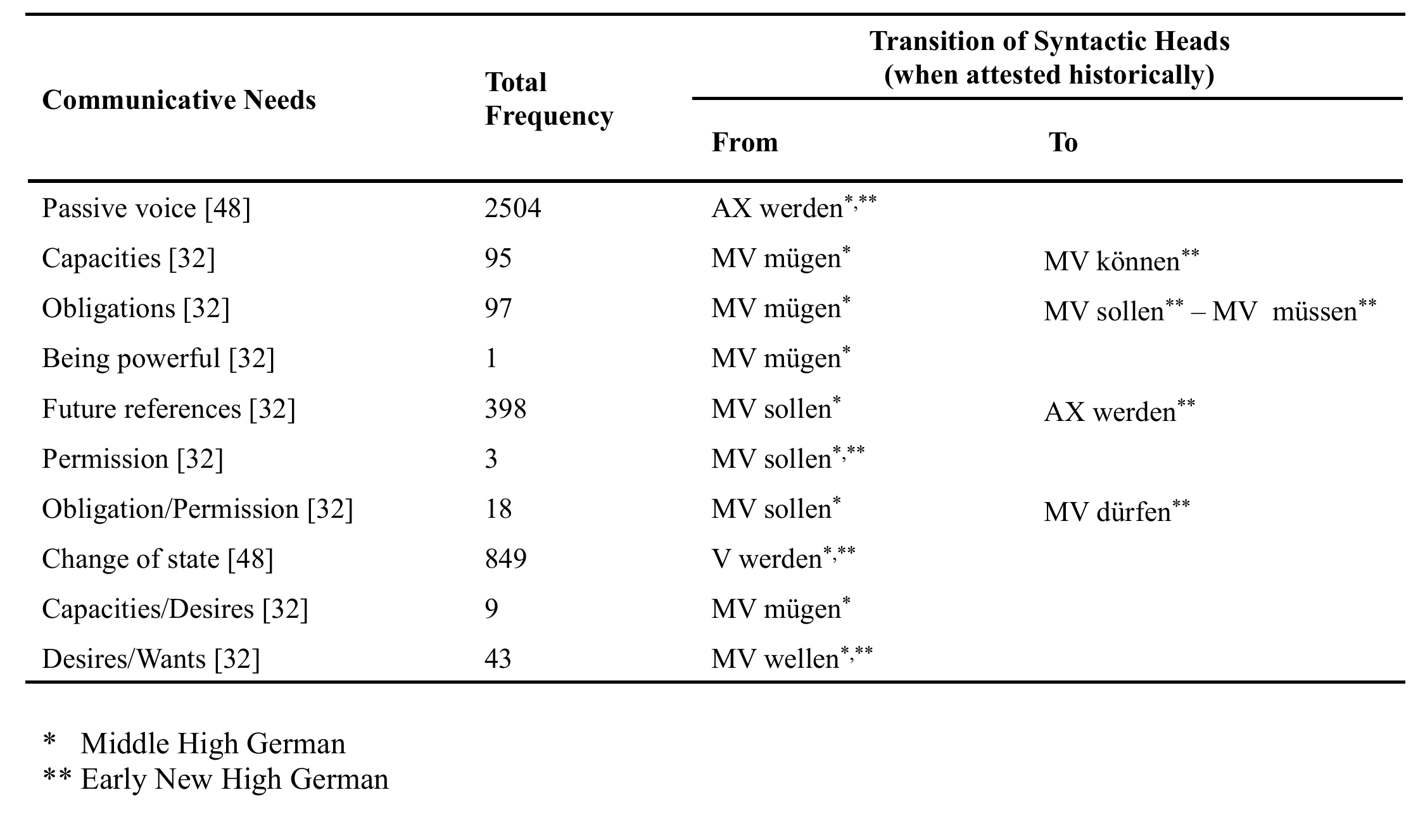}
    \label{tbl_communictive_needs}
\end{table}

\blankpage


\appendix

\renewcommand{\thesection}{\arabic{section}}
\renewcommand{\thesubsection}{\arabic{section}.\arabic{subsection}}

\renewcommand{\thefigure}{\arabic{figure}}
\setcounter{figure}{0}

\renewcommand{\thetable}{\arabic{table}}
\setcounter{table}{0}

\renewcommand\figurename{Figure S}
\makeatletter
\def\fnum@figure{\figurename\thefigure}
\makeatother

\renewcommand\tablename{Table S}
\makeatletter
\def\fnum@table{\tablename\thetable}
\makeatother

\begin{bibunit}

\setcounter{page}{1}
\resetlinenumber

\section*{SUPPLEMENTARY INFORMATION}

\begin{center}
\LARGE{The Hierarchical Organization of Syntax}
\end{center}

\begin{center}
    Babak Ravandi$^{1,2,*}$ and Valentina Concu$^{3},$\footnote{Authors equally contributed.}
    \\
    (bk.ravandi@gmail.com and vconcu@uninorte.edu.co)
    \\
    $^1${\small Network Science Institute, Northeastern University, Boston, USA}
    \vspace{-7pt}\\
    $^2${\small Department of Physics, Northeastern University, Boston, USA} 
    \vspace{-7pt}\\
    $^3${\small Department of Foreign Languages, Universidad del Norte, Barranquilla, Colombia}
\end{center}

\tableofcontents

\blankpage

\section{Dependency Syntactic Tree}

{Syntactic trees map the relationships between words in a sentence and are based on the notion of ``grammatical relation.'' According to this notion, the relationships between a head and a dependent are established in a given sentence. The syntactic tree below for the Modern English sentence ``I prefer the morning flight to Denver'' shows how such relationships are mapped.}

\begin{figure}[h!]
    \centering
    \includegraphics[scale=0.4]{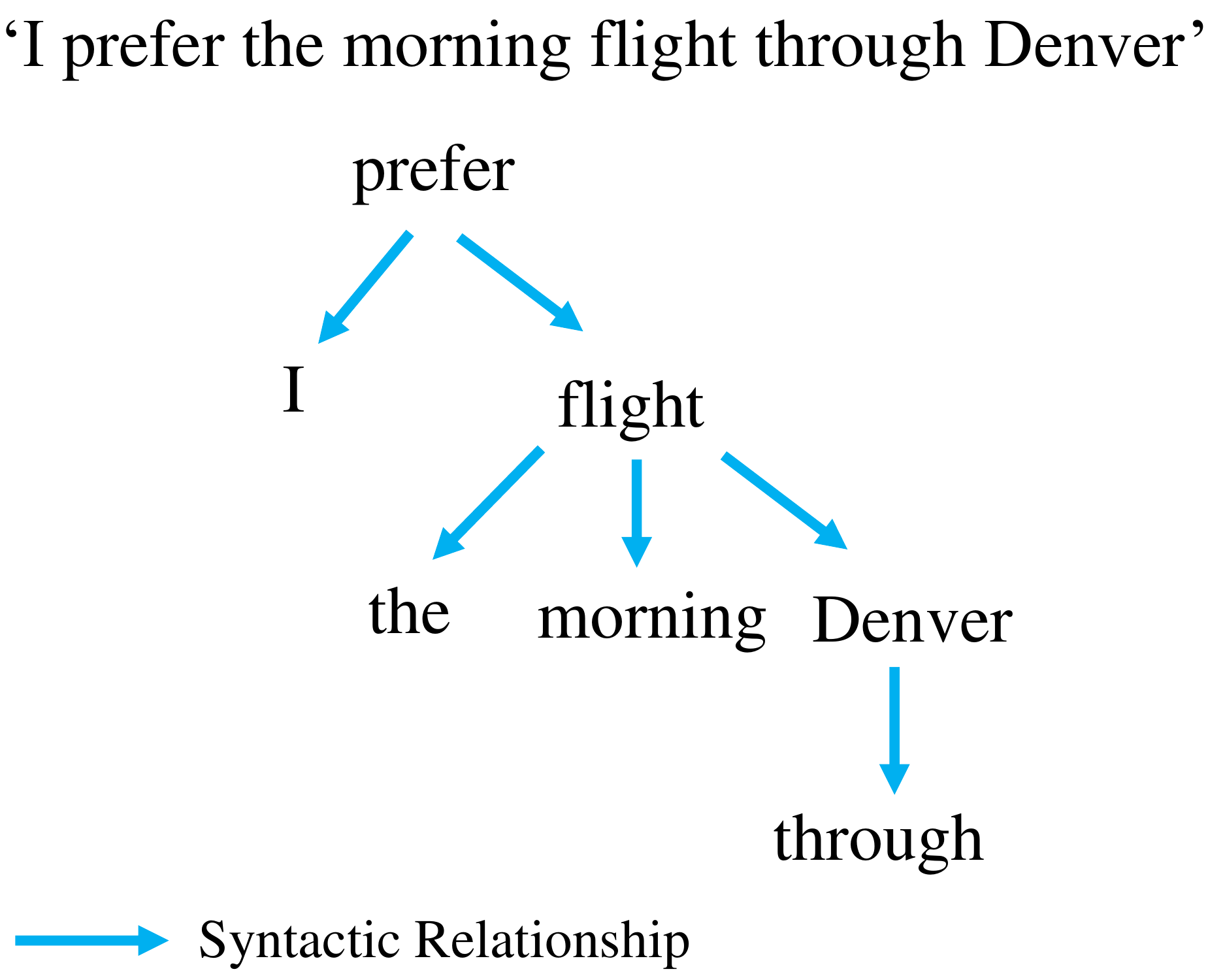}
    \vspace{10px}
    \caption{}
    \label{fig_syntactic_tree}
\end{figure}

\begin{figure}[h!]
    \contcaption{
\textbf{Dependency Syntactic Tree for ``I prefer the morning flight to Denver.''} 
 
The root is selected based on the concept of a governing element that corresponds to the most grammatically important element among the constituents. 
The governing element in this sentence is the verb ``prefer'' on which both the pronoun ``I'' and the noun ``flight'' depend. 
At the same time, the noun ``flight'' functions as the head with three dependents: the definite article ``the'', and the nouns ``morning'' and ``Denver''. 
``Denver'' has also another dependent, namely the preposition ``through 
(modified and used with permission from \cite{parsing2009speech}).
    }
\end{figure}

\section{Functions of \textit {Werden} in Modern German}

\textit {Werden} in Modern German can carry out several the grammatical functions: full verb, auxiliary of the passive voice, and auxiliary for the future tense. The following figure shows instances of \textit {werden} as it is used today:
\begin{figure}[h!]
    \centering
    \includegraphics[scale=0.50]{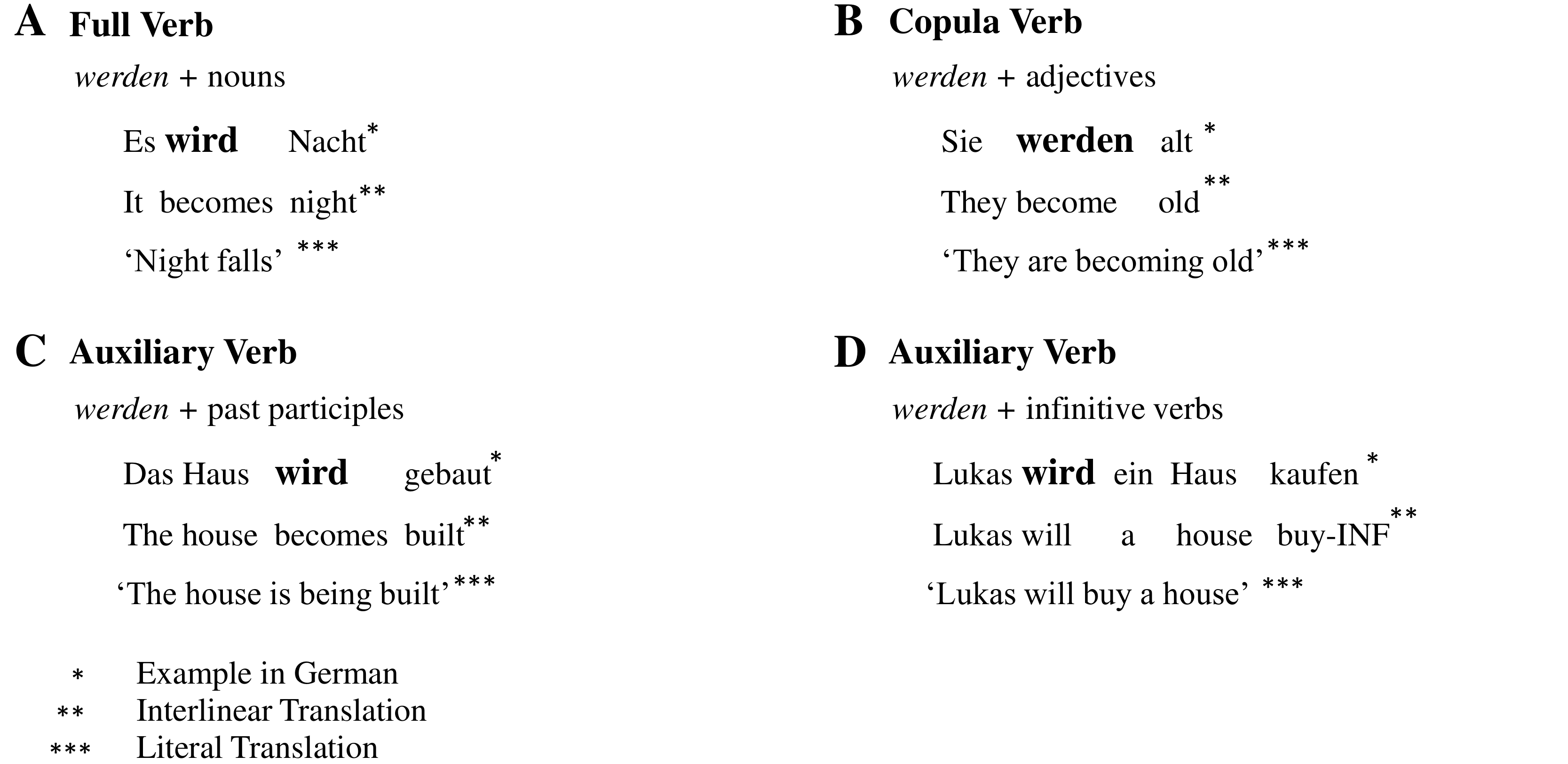}
    \caption{}
     \label{fig_werden_in_modern_german}
\end{figure}

\begin{figure}[h!]
    \contcaption{
\textbf{Functions of \textit{Werden} in Modern German.} 
        \textit {Werden} can be combined with a large variety of elements depending of the grammatical functions it is carrying out in a given sentence. 
        \textit {Werden} acquired the capacity to carry out these functions due to a process of \textit {Desemantisierung} (desemantization) that allowed this verb to appear in combinations with an increased type of elements throughout the centuries. 
        In the  examples \textbf{A-D}, the first line represents a sentence in Modern German. 
        The second line contains the interlinear morpheme-by-morpheme glosses (translation of each word). 
        The third line is the literal translation of the sentence in English. 
        The functions of \textit {werden} are: \textbf{(A) Full Verb} with its own meaning of `to become';
        \textbf{(B) Copula Verb} with nouns and adjectives (connecting a subject with a subject complement);
        %
        \textbf{(C-D) Auxiliary Verb} with past participles and infinitive verbs.
    }
\end{figure}


\blankpage

\section{Missing Annotations in the Middle and Early New High German Corpora}

The Middle and Early New High German corpora use different levels of annotations providing translation, grammatical functions, inflections, and more. However, in some cases, annotations that were vital to identify the function of \textit{werden} were missing:

\begin{figure} [h!]
    \includegraphics[scale=0.50]{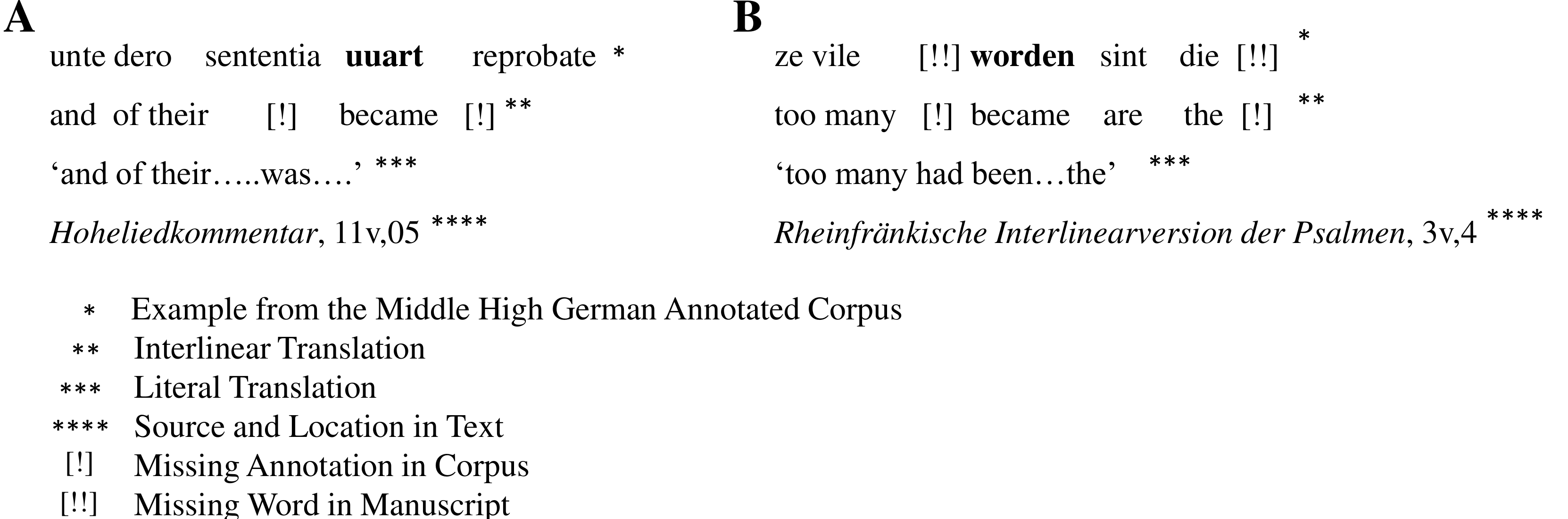}
    \vspace{5px}
     \caption{}

\textbf{Missing Words and Annotations in the Corpora.}

\textbf{(A)} The verb \textit{werden} (`\textit{uuart}') is used in combination with the words \textit{`sententia'}`sententia' and `\textit{rebrobate}', but no annotations are provided for any of them. 
Hence, it is not possible to establish how the target verb was used in this instance. 
Therefore, we filtered all sentences with this condition in our selection. 
\textbf{(B)} The missing words and annotations in this instance refer to elements not directly connected with \textit {werden}. The presence of the auxiliary verb \textit {sein} (to be - `\textit{sint}') \textit {werden} in the participle form (`\textit{worden}'), and the definite article \textit {die} suggest that the target verb has been used as a copula and can be, therefore, included in our selection.   
Accordingly, we included sentences with this type of missing annotations in our data as they do not interfere with identifying the grammatical function of \textit{werden}.

    
     \label{fig_werden_corpus_annotations}
\end{figure}

\blankpage

\section{Distributions of Hierarchical Levels}
\label{SI_sec_distribution_hierarchical_levels}

In this study, we used a local metric, the forward hierarchical level [26], as it mimics the top-down flow of our ASNs, which are all governed by heads that can be classified as vertices. The following figure shows that the distribution of the backward and forward hierarchical levels in ASNs:

\begin{figure}[h!]
    \
    \includegraphics[scale=0.5,page=1]{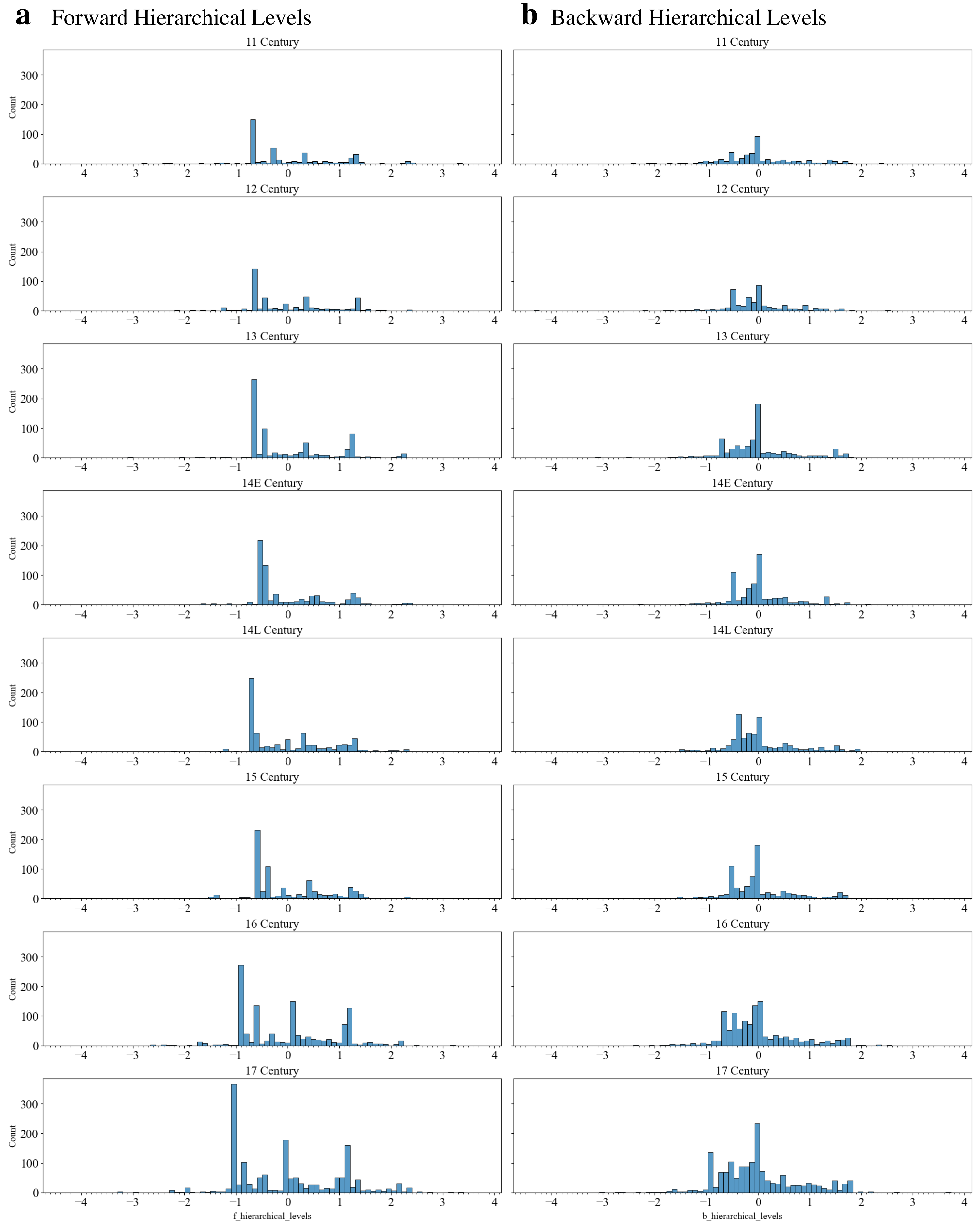}
    \caption{
\textbf{Distributions of forward and backward hierarchical levels.
}
    }
    \label{fig_hierarchical_levels_dist}
\end{figure}





\blankpage


\section{Potential Connections with Political Science}
\label{SI_sec_possible_connection_with_Zipfs}

\begin{figure}[b!]
  \centering
  \includegraphics[scale=0.24]{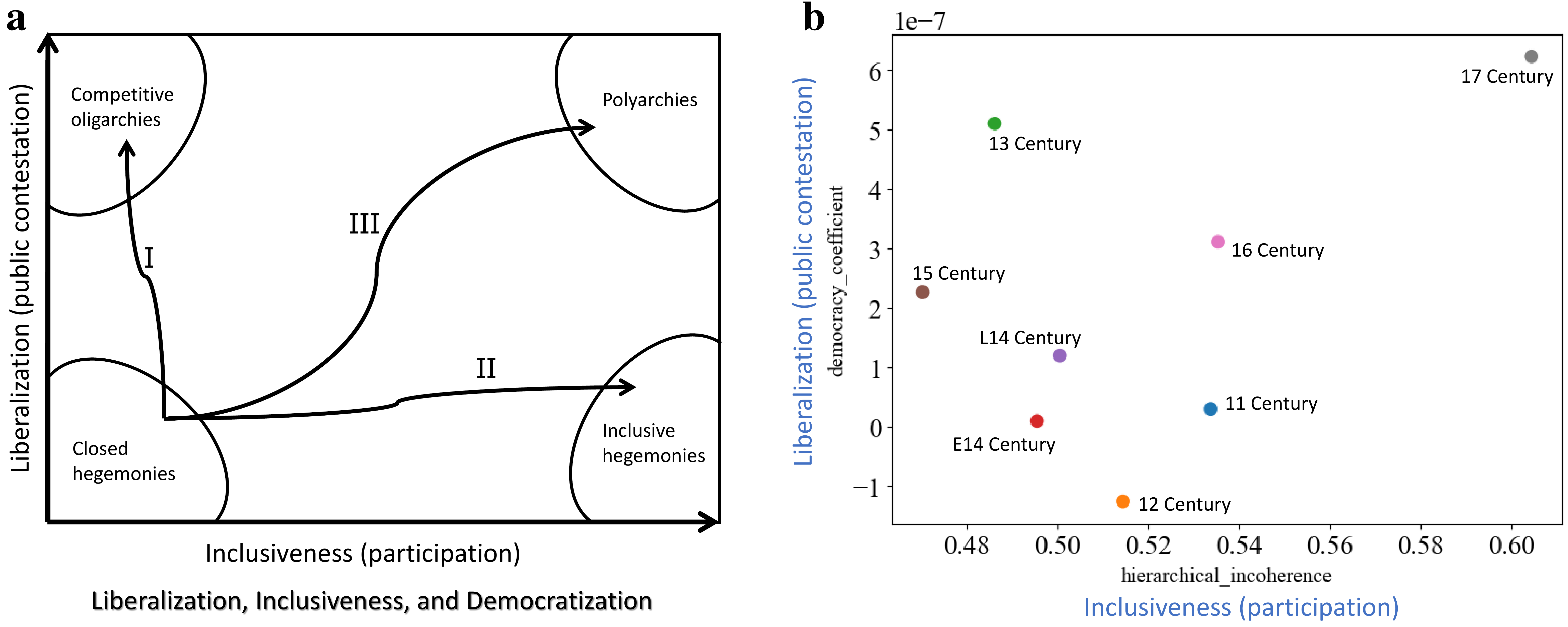}
  \vspace{5px}
  \caption{
\textbf{Trajectory of Language Change.} 
(\textbf{a}) Dahl indicates democratization of societies increase inclusiveness and public contestation, hence moving toward polyarchies.  
(\textbf{b}) The trajectory of language change with respect to macro-level hierarchical characteristics of ASNs signals that the same processes behind democratization may also be behind the change in language. 
  }
  \label{fig_phasespace}
\end{figure}

We observe that our ASNs have multiple heads, revealing a multi-hierarchical organization. Another term used to describe an entity with multiple hierarchies is polyarchy.  
Polyarchy has been defined in political science as a type of modern democratic representative government for large scale states or countries and means ``rule by the many'' \cite{dahl2020democracy}.
Polyarchy itself can be quantified through two measurements: liberalization and inclusiveness \cite{polyarchy_dahl_book}.  
Liberalization corresponds to public contestation, which translates into the degree by which a government allows any type of political opposition. 
Inclusiveness accounts for the ability of citizen to participate in controlling and contesting the conduct of the government. 
The more a society allows public contestation and the participation of its citizens to its political life, the more polyarchal will be, as illustrated in Figure S\ref{fig_phasespace}a.
Since languages and societies have co-evolved and affected each others throughout the centuries \cite{hopper2003grammaticalization}, we suspect that those processes that are involved in the democratization of a society could also be behind the language change. 
If so, these processes could provide us a conceptual framework to understand the trajectory of language evolution. 
Hence, we mapped liberalization and inclusiveness to democracy coefficient and hierarchical incoherence respectively.



The democracy coefficient of complex networks indicates the degree by which the influencers at the top of the hierarchies influence each other. 
In our ASNs, it translates into the ability of syntactic communicative hierarchies to share the expression of multiple communicative needs. 
The hierarchical incoherence indicates how neatly a graph can be partitioned into discrete levels, and it is derived from the hierarchical levels of nodes. 
Although the number of words in our texts does not vary much between centuries (the only exceptions are the 11th and the 12th centuries due to the scarcities of sources) the distribution of forward hierarchical levels tend to form more and more distinct strata of words as the language evolves (Figure S\ref{fig_hierarchical_levels_dist}). 
This signals that the functions of words in language are becoming more specific and, as a consequence, they are acquiring a more defined place in the syntactic communicative hierarchies. 
Figure S\ref{fig_phasespace} presents the trajectory of democracy coefficient and hierarchical incoherence of ASNs, we observe that both measures tend to increase as the language evolved. 


 As human beings started to settle down and organized themselves in fixed communities, the number of the members of these communities also increases. It is at this point that the first social hierarchical structures emerged \cite{dahl2020democracy}.
We can assume that the hierarchical structures that are found in ASNs also emerged as the number for words available to the speakers also increased. 
As already mentioned, Dahl claims that polyarchal democracies are democratic government on the large scale of the nation-state country and are the most effective way to govern, allowing to provide political representations as the number of citizens grows in a country \cite{dahl2020democracy}. 
In the same way, the multi-hierarchical structure of the language that we observed in our ASNs could be the most effective way for the language to handle the always increasing number of words and maintain its capability to satisfy the communicative needs of the speakers.

\putbib

\end{bibunit}

\end{document}